\documentclass[10pt,twocolumn,letterpaper]{article}

\usepackage{wacv}
\usepackage{times}
\usepackage{epsfig}
\usepackage{graphicx}
\usepackage{amsmath}
\usepackage{amssymb}

\usepackage{color}

\usepackage[pagebackref=true,breaklinks=true,letterpaper=true,colorlinks,bookmarks=false]{hyperref}

\wacvfinalcopy 

\def\wacvPaperID{152} 

\ifwacvfinal\fi
\newcommand{\comment}[1]{}

\newcommand{\reals}[1]{\mathbb{R}^{#1}}
\newcommand{\frameset}{\mathcal{F}}

\newcommand{\df}{F}
\newcommand{\norm}[1]{\left\|{#1}\right\|}

\begin{document}

\title{Ordered Pooling of Optical Flow Sequences for Action Recognition}

\author{
Jue Wang${^{1,3}}$ \hspace{2cm} Anoop Cherian${^{2,3}}$ \hspace{2cm} Fatih Porikli${^{1,2,3}}$ \\
 ${^1}$Data61/CSIRO, ${^2}$ Australian Center for Robotic Vision\\
${^3}$ Australian National University, Canberra, Australia
\\
{\tt\small jue.wang@anu.edu.au} \hspace{1cm}
{\tt\small anoop.cherian@anu.edu.au} \hspace{1cm}
{\tt\small fatih.porikli@anu.edu.au} \hspace{1cm}
}

\maketitle
\ifwacvfinal\thispagestyle{empty}\fi

\begin{abstract}
Training of Convolutional Neural Networks (CNNs) on long video sequences is computationally expensive due to the substantial memory requirements and the massive number of parameters that deep architectures demand. Early fusion of video frames is thus a standard technique, in which several consecutive frames are first agglomerated into a compact representation, and then fed into the CNN as an input sample. For this purpose, a summarization approach that represents a set of consecutive RGB frames by a single dynamic image to capture pixel dynamics is proposed recently. In this paper, we introduce a novel ordered representation of consecutive optical flow frames as an alternative and argue that this representation captures the action dynamics more effectively than RGB frames. We provide intuitions on why such a representation is better for action recognition. We validate our claims on standard benchmark datasets and demonstrate that using summaries of flow images lead to significant improvements over RGB frames while achieving accuracy comparable to the state-of-the-art on UCF101 and HMDB datasets. 

\end{abstract}

\section{Introduction}

Automatically recognizing human actions in videos is a challenging task since actions in real-world are often very complex, may involve hard to detect objects and tools, may have different temporal speeds, can be contaminated by action clutter, or the same action can vary significantly from one actor to another, among several other factors. Nevertheless, developing efficient solutions for this problem could facilitate and nurture many applications, including visual surveillance, augmented reality, video summarization, and human-robot interaction. The recent resurgence of deep learning has demonstrated a significant promise in ameliorating the difficulties in recognizing actions. However, such solutions are still far from being practically useful, and thus this problem continues to be a popular research topic in computer vision~\cite{aggarwal2011human,feichtenhofer2016convolutional,poppe2010survey,simonyan2014two,wang2013action,wang2015action}.

\begin{figure}[t]
\begin{center}
   \includegraphics[width=1\linewidth]{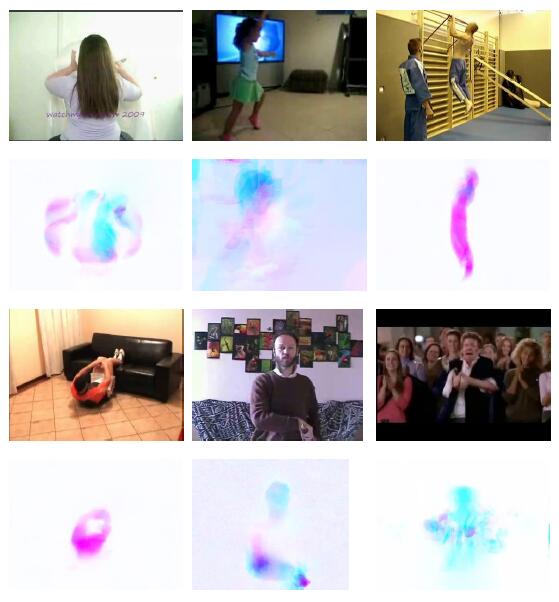}
\end{center}
   \caption{Examples of dynamic flow image and related RGB video frames. From left to right and top to bottom, the action classes are: ``brush hair'', ``cart wheel'', ``pull up'', ``sit up'', ``draw sword'' and ``clip''. Each dynamic flow image is a two channel image that compactly summarizes a sequence of optical flow frames from a video sequence.}
\label{fig:1}
\end{figure}

Usually, deep architectures for video based action recognition take as input short video clips consisting of one to a few tens of frames. Using longer subsequences would require deeper networks or involve a huge number of parameters, which might not fit in the GPU memory, or may be problematic to train due to computational complexity. This restriction and thus clipping of the temporal receptive fields of the videos to short durations prohibit CNNs from learning long-term temporal evolution of the actions, which is very important in recognition especially when the actions are complex. 


One standard way to tackle this difficulty in capturing long-term dependencies is to use temporal pooling that can be applied either when providing the input to the CNN or after extracting features from intermediate CNN layers~\cite{bilen2016dynamic}. 

In this paper, we explore a recently introduced early fusion scheme based on a Ranking SVM~\cite{fernando2015modeling} formulation where several consecutive RGB frames in the video are fused to generate one ``dynamic image'' by minimizing a quadratic objective with temporal order constraints. The main idea of this scheme is that the parameters learned after solving this formulation capture the temporal ordering of the pixels from frame-to-frame, thus summarizing the underlying action dynamics. 


One drawback of the rank pooling approach is that the formulation does not directly capture the motion associated with the action -- it only captures how the pixel RGB values change from frame-to-frame. Typically, the video pixels can be noisy. Given that the pooling constraints in this scheme only look at increasing pixel intensities from frame-to-frame, it can fit to noise pixels that adhere to this order, however, are unrelated to the action. To mitigate these issues, in this paper, we look at the rank pooling formulation in the context of optical flow images instead of RGB frames.


It is clear that optical flow can easily circumvent the above problems associated with sequences of RGB frames. The flow by itself captures the objects in motion in the videos, and thereby capturing the action dynamics directly, while thresholding the flow vectors helps to avoid noise. Thus, we posit that summarizing sequences of optical flow can be significantly more beneficial than using RGB frames for action recognition. By solving the rank-SVM formulation (Section~\ref{method}), we generate flow images that summarize the action dynamics, dubbed~\emph{dynamic flow images}. These images are then used as input to a standard action recognition CNN architecture to train for the actions in the sequences. In Figure~\ref{fig:1}, we show a few sample dynamic flow images generated using the proposed technique for the respective RGB frames. 

A natural question here can be regarding the intuitive benefit of using such a flow summarization scheme, given that standard action recognition CNN frameworks already use a stack of flow frames. Note that such flow stacks usually use only a few frames (usually 10), while using the dynamic flow images, we summarize several tens of flow frames, thereby capturing long-term temporal context. 


To validate our claims, we provide extensive experimental comparisons (Section~\ref{experiment}) on two standard benchmark action recognition datasets, namely (i) the HMDB-51 dataset, and (ii) the UCF-101 dataset. Our results show that using dynamic flow images lead to significant improvements in action recognition performance. We find that this leads to 4\% improvement on HMDB-51\cite{kuehne2011hmdb} and 6\% on the UCF-101\cite{soomro2012ucf101} dataset in comparison to using dynamic RGB images without combining any other methods.


Before moving on to explaining our scheme in detail, we summarize below our important contributions. 
\begin{enumerate}
\item We provide an efficient and powerful video representation,~\emph{dynamic flow image}, generated based on the Ranking SVM formulation. Our representation aggregates local action dynamics (as captured by optical flow) over subseqeunces while preserving the temporal evolution of these dynamics. 

\item We provide an action recognition framework based on the popular two-stream network \cite{simonyan2014two}, that also combines dynamic flow images, RGB frames (for action context), and hand-crafted trajectory features.

\item We provide extensive experimental comparisons demonstrating the effectiveness of our scheme on two challenging benchmark datasets.
\comment{
able to compress the original dataset by a compression factor, which make the dataset relatively small and is still compatible for training CNNs. 3. Dynamic flow summarize the action dynamic inside the video, which has significant improvement compared with the \cite{bilen2016dynamic} and we propose a multi-stream algorithm combined with dynamic flow image, which make fully comparesion with \cite{bilen2016dynamic} in two most public datasets, UCF101 \cite{soomro2012ucf101} and HMDB51\cite{kuehne2011hmdb}. 4.
}
\end{enumerate}

\section{Related Work}\label{relatedwork}
Extracting discriminative features from videos is the first and most important step in action recognition. Such features can be roughly divided into two groups: handcrafted local features and deep learned features. Of course, there are also many interesting methods that are not limited to these, such as using genetic programming to learn spatial-temporal features automatically ~\cite{liu2016learning}, using human shape evolution of skeletons ~\cite{amor2016action}, and so on. However, we will limit our survey to the techniques most related to the approach presented in this paper.

\noindent\textbf{Handcrafted Local Features. } Using local features to generate representations for action recognition is popular because they make the recognition robust to noise, background motion, or illumination changes. One noteworthy such representation is described in~\cite{wang2013action}, where dense trajectories, that capture the dynamics of actions, is used to define regions of interest in the video. Local features (such as HOG, HOF, MBH, etc.), that directly relate to the action can then be extracted from these regions to train classifiers.  There have been extensions of this approach to also include semantics of the action, such as using Fisher vectors~\cite{sadanand2012action,wu2015good}, using bag of visual words model ~\cite{peng2016bag}, combining with depth data ~\cite{liu2016hierarchical}, or using human pose~\cite{tensor_eccv}. However, the recent trend is towards automatically learning useful features in a data-driven way via convolutional neural networks~\cite{bilen2016dynamic, feichtenhofer2016convolutional,wang2015action}, and we follow this trend in this paper.

\noindent\textbf{Deep Learned Features. } Deep learning methods have been extensively used for computer vision tasks since the work of Krizhevsky et al.~\cite{krizhevsky2012imagenet} for object recognition. There have been extensions of this model for the problem of action recognition recently. In~\cite{simonyan2014two}, a two-stream CNN model is proposed for this task with very promising results. The two-stream model is enhanced using a VGG network in~\cite{feichtenhofer2016convolutional}, and intermediate fusion of the convolutional layers incorporated. A trajectory-pooled CNN setup is described in~\cite{wang2015action} that fuses CNN features along action trajectories. Early and late fusion of CNN feature maps is proposed in~\cite{karpathy2014large,yue2015beyond}. Almost all these methods combine the feature maps without accounting for the temporal evolution of these features; as a result, might lose capturing temporal action dynamics effectively.

To account for this shortcoming, Fernando et al.~\cite{fernando2015modeling} proposed a pooling formulation on video features that accounts for the temporal order of frames. This formulation  is extended for RGB frames in~\cite{bilen2016dynamic,fernando2016discriminative}. We base this paper on this extension, however deviate from their approach in that we use optical flow frames instead of RGB frames for capturing the action dynamics. To the best of our knowledge, we are not aware of any other work that has analyzed the advantages of using optical flow images in the rank-SVM formulation for action recognition. We showcase extensive experiments to substantiate the benefits against using RGB frames. 

We also note that there have been several other deep learning models devised for action modeling such as using recurrent neural networks~\cite{baccouche2011sequential,du2015hierarchical} and long-short term memory networks~\cite{donahue2014long,li2016action,yue2015beyond}. While, we acknowledge that using recurrent architectures may benefit action recognition, usually datasets for this task are often small or are noisy, that training such models will be challenging. Thus, we focus on advancing action representations in this paper for effective CNN training.

\section{Proposed Method}\label{method}

In this section, we introduce our dynamic flow images; we describe the necessary formulations to generate these images, followed by an exposition to a multi-stream CNN framework for action recognition. Note that our dynamic flow formulation is inspired by the recently introduced dynamic image framework~\cite{bilen2016dynamic}, however, our contribution is to show the shortcomings of that formulation and propose remedies.

\comment{
In this section, we describe the theory of dynamic flow image. Similar as the \cite{bilen2016dynamic}, dynamic flow image is a standard three-channel image that contain the dynamic information over a video sequence. And one video sequence could be represented by one or multiple dynamic flow images. After that, how to use such dynamic flow image to train CNN is presented. At last, we show a multi-stream algorithm and introduce the fusion method.
}

\begin{figure*}
\begin{center}
	\includegraphics[width=1\linewidth]{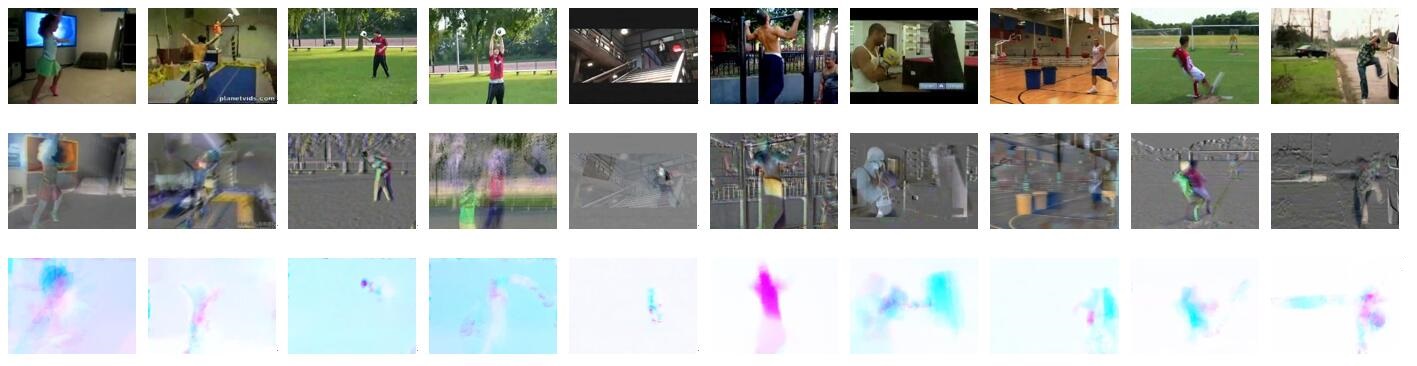}
\end{center}
   \caption{Qualitative comparisons between dynamic images~\cite{bilen2016dynamic} (second row) and our proposed dynamic flow image (third row) representation. An RGB frame from the respective video is also shown (first row). As can be seen from the visualizations, dynamic flow focuses on actions while dynamic images can be contaminated by background pixels.}
\label{fig:2}
\end{figure*}

\subsection{Formulation}
Let us assume that we are provided with a sequence of $n$ consecutive optical flow images $\frameset=\left[ f_1, f_2,\cdots, f_n\right]$, where each $f_i\in\reals{d_1\times d_2\times 2}$, where $d_1, d_2$ are the height and width of the image. We assume a two-channel flow image corresponding to the horizontal and vertical components of the flow vector, which we denote by $f^u_i$ and $f^v_i$ respectively. Our goal is to generate a single "dynamic flow" image $\df\in\reals{d_1\times d_2\times 2}$ that captures the temporal order of the flow images in $\frameset$. We use the following formulation to obtain this representation.
\begin{align}
\label{eq:2}
&\min_{\df\in\reals{d_1\times d_2\times 2}, \xi\geq 0} \quad \norm{\df}^2 + C\sum_{i<j} \xi_{ij}\\
\text{s. t. }& \langle \df, f_i \rangle \leq \langle \df, f_j \rangle - 1 + \xi_{ij},\quad \forall i<j,\nonumber
\end{align}
where $\langle .,.\rangle$ represents an inner product between the original flow vectors and the dynamic flow image to be found. As one may recall, this formulation is similar to the Ranking-SVM formulation~\cite{joachims2002optimizing}, where the inner product captures the ranking order, which in our case corresponds to the temporal order of the frames in the sequence. While, ideally, we enforce that the projection (via the inner-product) of the flow frames to the dynamic flow image is lower bounded by one, we relax this constraint using the variables $\xi$ as is usually done in a max-margin framework. While, the above formulation is a direct adaptation of the Ranking SVM formulation, there are a few subtleties that need to be taken care when using this formulation for optical flow images. We will discuss these issues next. 

It is often observed that a direct use of raw optical flow in~\eqref{eq:2} is often inadequate. This is because computing optical flow is a computationally expensive and difficult task, and the flow solvers are often prone to local minima, leading to inaccurate flow estimates. In order to circumvent such issues, we instead apply the algorithm on flow images computed over running averages; such a scheme averages the noise in the flow estimates. For the flow set $\frameset$, we represent the resulting smoothed flow image for frame $f_t$ as:
\begin{equation}
\hat{f}_t= \frac{1}{t}\sum_{i=1}^t f_i.
\label{eq:3}
\end{equation}
\noindent We compute the dynamic flow on such smoothed flow images. Note that since flow is signed, such averaging will cancel white noise.

Another issue specific to optical flow images is that the two flow channels $u$ and $v$ are coupled. They are no more the color channels as in an RGB image, instead they represent velocity vectors, which together  capture the motion vector at a pixel. However, our ranking formulation assumes indepedence in the channels. One way to avoid this difficulty is to decorrelate these channels via diagonalization; i.e., use the singular vectors associated with a short set of flow images as representatives of the original flow images. However, our experiments showed that this did not lead to significant improvements. Thus, we choose to ignore this coupling in the current paper, and assume each channel is independent when creating the dynamic flow images. Our experiments (in Section~\ref{experiment}) show that this assumption is not detrimental.

Incorporating the above assumptions into the objective, and assuming $F_u,F_v\in\reals{d_1\times d_2}$ are the two dynamic flow channels corresponding to the horizontal and vertical flow directions, we can rewrite~\eqref{eq:2} as:
\begin{align}
\min_{F_u,F_v\in\reals{d_1\times d_2}, \xi\geq 0} & \norm{F_u}^2 + \norm{F_v}^2 + C\sum_{i<j} \xi_{ij}\nonumber\\
\text{ subject to}\quad\quad\quad &\nonumber\\
\langle F_u, \hat{f}^u_i \rangle + \langle F_v, \hat{f}^v_i \rangle &\leq 
\langle F_u, \hat{f}^u_j \rangle + \langle F_v, \hat{f}^v_j \rangle -1 + \xi_{ij}, \forall i<j.
\end{align}

We solve the above optimization problem using the libSVM package as described in~\cite{bilen2016dynamic}. Once the two flow images $F_u$ and $F_v$ are created, they are stacked to generate a two channel dynamic flow image, which is then fed to a multi-stream CNN as described for learning actions (as described in the next section). In Figure~\ref{fig:2}, we show several examples of the dynamic flow images, and their respective RGB frames, and dynamic RGB images. As is clear from these visualizations, the dynamic flow images summarizes the actual action dynamics in the sequences, while the dynamic RGB images include averaged background pixels and thus are may include dynamics unrelated to recognizing actions.

\subsection{Three-Stream Prediction Framework}
\begin{figure*}[ht]
\centering
	\includegraphics[width=0.95\linewidth]{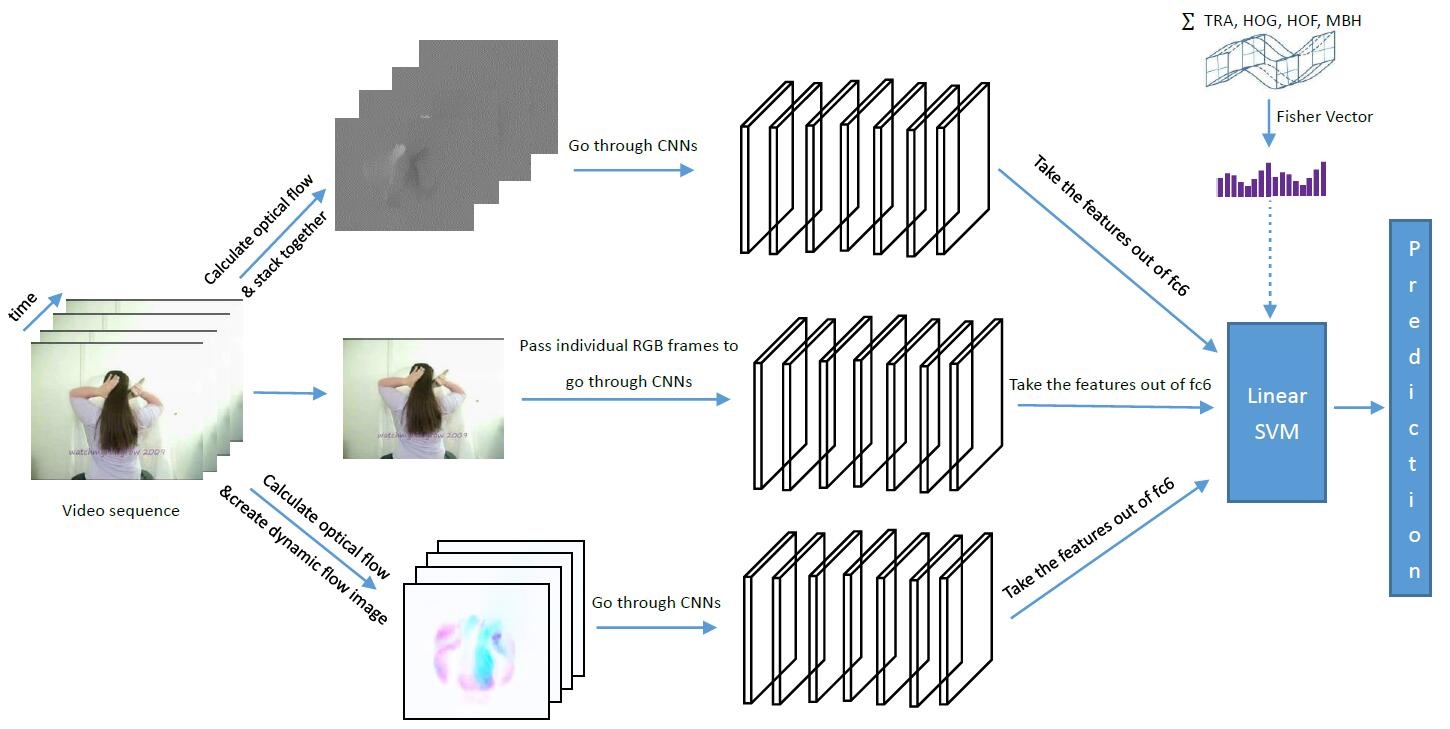}
   \caption{Architecture of our dynamic-flow CNN based classification setup. Our three-stream CNN consists of an optical flow stream taking stacks of flow frames, an RGB stream taking single RGB frames, and our dynamic flow stream taking single images, each image summarizing the action dynamics over 25 flow frames.}
\label{fig:3}
\end{figure*}

Next, we propose a three-stream CNN setup for action recognition. This architecture is an extension of the popular two-stream model~\cite{simonyan2014two} that takes as input individual RGB frames in one stream and a small stack of optical flow frames (about 10 flow images) in the other. One shortcoming of this model is that it cannot see long-range action evolution, for which we propose to use our dynamic flow images (that summarizes about 25 flow frames in one dynamic flow image). As is clear, each such stream looks at complementary action cues. Our overall framework is illustrated in Figure~\ref{fig:3}.

To be precise, for the dynamic flow stream, for each video sequence, we generate multiple dynamic flow images. In order to achieve this, we first split the input flow video into several sub-sequences each of length $w$ and generated at a temporal stride $s$. For each sub-sequence, we construct a dynamic flow image using the optical flow images in this window. We associate the same ground truth action label for all the sub-sequences, thus effectively increasing the number of training videos by a factor of $\frac{n}{s}$, where $n$ is the average number of frames in the sequences. Note that we use a separate CNN stream on dynamic flow images. Given that action recognition datasets are usually tiny, in comparison to image datasets (such as ImageNet), increasing the training set is usually necessary for the effective training of the network.

\subsection{Practical Extensions}
We use the TVL1 optical flow~\cite{zach2007duality} algorithm to generate the flow images using its OpenCV implementation. For every flow image, we subtract the median flow, thus removing camera motion if any (assuming flow from the action dynamics occupies a small fraction with respect to the background). The resulting flow images are then thresholded in the range of $[-20, 20]$ pixels and setting every other flow vector to zero. This step thus removes unwanted flow vectors, that might correspond to noise in the images. Next, we scale the flow to the discrete range of $[0,255]$, and convert each flow channel as a gray-scale image. A set of such transformed flow images are then used as input to the rank-SVM formulation in~\eqref{eq:3}, thereby generating one dynamic flow image per sub-sequence. Using libSVM package, it takes about 0.25 seconds to generate one dynamic flow image on a single core CPU combining 25 flow frames each of size $224\times 224\times 2$.

\comment{
Alternatively, we could just produce one dynamic flow image for each sequence, which summarize the entire optical flow of one video into one single dynamic flow image. However, \cite{bilen2016dynamic} has proved in their experiment that the performance of this method is much worse than the previous one. The reason can be concluded into two perspective. First, even though fine-tuning CNNs does not require as much data as training CNNs from scratch, adequate training data is necessary.  Most  of the popular dataset at this stage is relatively small. If just generate one dynamic image to represent one video, there are only a few training samples, which is far less than what is required and will be hard to avoid over-fitting. Another reason is that, each video has different length, which means each dynamic image are going to contain different amount of dynamic information. This would be a challenge for CNNs to learn, because some training samples are too noisy and some others  are not.

In order to figure how much valid dynamic information each dynamic flow image can carry, we experiment different size of window when creating dynamic flow (see section 4).Through the performance of CNNs, we are able to learn how to build optimized dynamic flow. 
}

\section{Experiments}\label{experiment}
In this section, we first describe the datasets used in our experiments, followed by an exposition to the implementation details of our framework and comparisons to prior works.

\subsection{Datasets}
We use two standard datasets for the task: (i) the HMDB51 dataset~ \cite{kuehne2011hmdb} and (ii) the UCF101 dataset~\cite{soomro2012ucf101}. For both the datasets, the standard evaluation protocol is the average classification accuracy over three splits.

\noindent\paragraph{HMDB51 dataset \cite{kuehne2011hmdb}} consists of 6766 video clips distributed in 51 action categories. The videos are downloaded from Youtube and are generally of low resolution. 

\noindent\paragraph{UCF101 dataset~\cite{soomro2012ucf101}} is a relatively larger dataset, containing 13,320 videos spanning over 101 actions. The videos are mostly from sports activities and contain significant camera motions and people appearance variations.

\subsection{Implementation Details}
\textbf{Training CNNs.} In the experiments to follow, we use the two successful CNN architectures, namely Alexnet~\cite{krizhevsky2012imagenet} and VGG-16~\cite{simonyan2014very}. We use the Caffe toolbox \cite{jia2014caffe} for the implementation. As the number of training videos is substantially limited to train a standard deep network from scratch, we decided to fine-tune the networks from models pre-trained for image recognition tasks (on the ImageNet dataset). On the training subsets of our dataset, we fine-tune all the layers of the respective networks with an initial learning rate of $10^{-4}$ for VGG-16 and $10^{-3}$ for Alex net with a drop-out of 0.85 for 'fc7' and 0.9 for 'fc6' as recommended in~\cite{simonyan2014very,feichtenhofer2016convolutional}. The drop-out is subsequently increased when the validation loss begins to increase. The network is trained using SGD with a momentum of 0.9 and weight decay of 0.0005. We use a mini-batch size of 64 for HMDB51 and 128 for UCF101.

\noindent\paragraph{Fusion Strategy.} As alluded to earlier, we use a three-stream network with RGB, stacked-optical flow, and dynamic flow images. The three networks are trained separately. During testing, output of the fc6-layer of each stream is extracted (see Figure~\ref{fig:3}). These intermediate features are then concatenated into a single vector, which is then fused via a linear SVM. We also experimented with features from the fc7 layer, however we found them to perform slightly inferior  compared to fc6. As is typically done, we also extract dense trajectory features from the videos (such as HOG, HOF, and MBH)~\cite{wang2013dense}, which are encoded using Fisher vectors, and are concatenated with the CNN features before passing to the linear SVM.

\subsection{Experimental Results}
\begin{table}[]
\centering
\caption{Influence of using different number of consecutive flow frames (window size) to construct one dynamic flow image. This experiment uses the split 1 of HMDB51 dataset with the VGG-16 CNN model.}
\begin{tabular}{ll}
\hline
Dynamic Flow window size           & Accuracy \\ \hline
15		                           & 48.23\%  \\
25                                 & \textbf{48.75\%}  \\
30                                 & 46.60\%  \\ \hline
\end{tabular}
\label{table:1}
\end{table}
We organize our experiments into various categories, namely (i) to decide the hyper-parameters of our formulation (e.g., the window size to generate dynamic flow images), (ii) benefits of using dynamic flow images against dynamic RGB images and other data modalities, (iii) influence of the CNN architecture, (iv) complementarity of the dynamic flow images to hand-crafted features, and (v) comparisons against the state of the art. Below, we detail each of these experiments.

\noindent\paragraph{Hyper-parameters:} We tested the performance of dynamic flow images using different temporal window sizes, i.e., the number of consecutive flow frames used for generating one dynamic flow image. The results of this experiment on the HMDB51 dataset split 1 is provided in Table~\ref{table:1}. As is clear, increasing window sizes is not beneficial as it may lead to more action clutter. Motivated by this experiment, we use a window size of 25 at a temporal stride of 5 in all the experiments in the sequel. 

\begin{table}[]
\centering
\caption{Evaluation on HMDB51 split 1 using VGG-16 model.}
\begin{tabular}{ll}
\hline
Method                                         & Accuracy \\ \hline
Static RGB\cite{feichtenhofer2016convolutional}& 47.06\%  \\
Stacked Optical Flow \cite{feichtenhofer2016convolutional}         
                                               & 55.23\%  \\
Dynamic Image~\cite{bilen2016dynamic}                                  & 44.74\%  \\ 
Dynamic Flow                                   & 48.75\% \\
Dynamic Image+RGB                              & 47.96\%  \\
Dynamic Flow+RGB                               & \textbf{58.30\%}\\
Dynamic Image+Dynamic Flow+RGB                 & 54.86\% \\
(S)Optical Flow+RGB\cite{feichtenhofer2016convolutional}
											   & 58.17\% \\
Dynamic Image+RGB+(S)Optical Flow              & 55.40\%  \\
Dynamic Flow+RGB+(S)Optical Flow               & \textbf{61.70\%}\\\hline
\end{tabular}
\label{table:2}
\end{table}

\begin{table}[]
\centering
\caption{Evaluation on UCF101 using AlexNet CNN model.}
\begin{tabular}{ll}
\hline
Method                             & Accuracy \\ \hline
\multicolumn{2}{l}{On split 1}\\\hline
Static RGB\cite{wang2015action}                  & 71.20\%  \\
Stacked Optical Flow\cite{wang2015action}        & 80.10\%  \\
Dynamic Flow                                     & 75.36\%  \\
Dynamic Flow + RGB                               &\textbf{84.93\%}\\
(S)Optical Flow + RGB\cite{wang2015action}       & 84.70\%  \\ 
Dynamic Flow + RGB + (S)Optical Flow             & \textbf{88.63\%}\\ \hline
\multicolumn{2}{l}{Over three splits\footnote{The dynamic RGB image results using Alexnet on UCF101 three splits was taken directly from~\cite{bilen2016dynamic}.}}\\\hline
Static RGB                 & 70.10\%  \\ 
Dynamic Flow                                      & 76.19\%  \\
Dynamic Image\cite{bilen2016dynamic}									  & 70.90\%  \\
Dynamic Image + RGB\cite{bilen2016dynamic}        & 76.90\%  \\
Dynamic Flow + RGB                                & \textbf{84.93\%} \\ \hline
\end{tabular}
\label{table:3}
\end{table}

\noindent\paragraph{Benefits over Dynamic Images (DI):} As discussed in Section~\ref{method}, dynamic flow (DF) captures the action dynamics directly in comparison to the dynamic RGB images~\cite{bilen2016dynamic}. To demonstrate this, we show experiments using the VGG-16 network in Table~\ref{table:2} on the HMDB51 dataset and using the Alexnet network on the UCF101 dataset in Table~\ref{table:3}\footnote{The results of Alexnet on UCF101 was taken directly from~\cite{bilen2016dynamic}.}. We also show comparisons to RGB, stack of flow, and various combinations of them. As is clear from the two tables, DF + RGB is about 9-10\% better than DI + RGB consistently on both the datasets and both CNN architectures. Surprisingly, combining DI with DF + RGB leads to a reduction in performance of about 4\% (Table~\ref{table:2}). This, we suspect, is because DI includes pixel dynamics from the scene background that may be unrelated to the actions and may confuse the subsequent classifier; such noise is avoided by computing optical flow. We investigate this further, in Table~\ref{table:4}, where we provide the per-class accuracy on HMDB51 for DI+RGB, DF+RGB, and DI+DF+RGB. We find that 29 of the 51 actions in this dataset lose in performance when using DI+DF+RGB as against DF+RGB, which implies that DF+RGB is indeed a better representation for actions. 


\begin{table}[]
\centering
\caption{Accuracy on each class in HMDB51 split 1 using different methods.}
\begin{tabular}{llll}
Action          & DI+RGB & DF+RGB & DI+DF+RGB \\
\hline
brush\_hair     & 60\%   & \textbf{77\%}   & \textbf{77\%}      \\
cartwheel       & 3\%    & \textbf{23\%}   & 13\%      \\
catch           & 27\%   & \textbf{43\%}   & \textbf{43\%}      \\
chew            & 43\%   & \textbf{60\%}   & \textbf{60\%}      \\
clap            & 38\%   & \textbf{52\%}   & \textbf{52\%}      \\
climb\_stairs   & \textbf{60\%}   & 53\%   & \textbf{60\%}      \\
climb           & 67\%   & \textbf{73\%}   & \textbf{73\%}      \\
dive            & 50\%   & \textbf{60\%}   & 53\%      \\
draw\_sword     & 30\%   & \textbf{47\%}   & 40\%      \\
dribble         & 73\%   & \textbf{90\%}   & 80\%      \\
drink           & 20\%   & \textbf{43\%}   & \textbf{43\%}      \\
eat             & 33\%   & \textbf{50\%}   & 37\%      \\
fall\_floor     & 34\%   & 24\%   & \textbf{38\%}      \\
fencing         & 63\%   & \textbf{77\%}   & 70\%      \\
flic\_flac      & 30\%   & \textbf{50\%}   & \textbf{50\%}      \\
golf            & 93\%   & \textbf{97\%}   & 93\%      \\
handstand       & 50\%   & \textbf{70\%}   & 60\%      \\
hit             & 14\%   & 25\%   & \textbf{29\%}      \\
hug             & 53\%   & \textbf{67\%}   & 57\%      \\
jump            & 31\%   & \textbf{52\%}   & 41\%      \\
kick\_ball      & 37\%   & \textbf{40\%}   & 33\%      \\
kick            & 7\%    & \textbf{20\%}   & 17\%      \\
kiss            & 83\%   & 83\%   & \textbf{87\%}      \\
laugh           & 50\%   & \textbf{70\%}   & 57\%      \\
pick            & 30\%   & \textbf{40\%}   & 37\%      \\
pour            & 83\%   & \textbf{97\%}   & 93\%      \\
pullup          & 87\%   & \textbf{100\%}  & \textbf{100\%}     \\
punch           & 63\%   & 63\%   & \textbf{70\%}      \\
push            & 70\%   & \textbf{87\%}   & 77\%      \\
pushup          & 57\%   & 60\%   & \textbf{67\%}      \\
ride\_bike      & 93\%   & 93\%   & \textbf{97\%}      \\
ride\_horse     & 80\%   & \textbf{83\%}   & 77\%      \\
run             & 32\%   & \textbf{50\%}   & 39\%      \\
shake\_hands    & 70\%   & 73\%   & \textbf{77\%}      \\
shoot\_ball     & 80\%   & \textbf{87\%}   & 83\%      \\
shoot\_bow      & 87\%   & \textbf{93\%}   & 87\%      \\
shoot\_gun      & 67\%   & 60\%   & \textbf{70\%}      \\
sit             & 37\%   & \textbf{57\%}   & 53\%      \\
situp           & \textbf{87\%}   & 77\%   & 83\%      \\
smile           & 40\%   & \textbf{43\%}   & 40\%      \\
smoke           & 40\%   & 47\%   & \textbf{53\%}      \\
somersault      & 60\%   & \textbf{83\%}   & 70\%      \\
stand           & 20\%   & \textbf{23\%}   & 20\%      \\
swing\_baseball & 13\%   & \textbf{17\%}   & \textbf{17\%}      \\
sword\_exercise & 13\%   & \textbf{30\%}   & 17\%      \\
sword           & \textbf{31\%}   & 17\%   & 21\%      \\
talk            & 57\%   & \textbf{67\%}   & 60\%      \\
throw           & 10\%   & \textbf{33\%}   & 7\%       \\
turn            & 37\%   & \textbf{57\%}   & 47\%      \\
walk            & 40\%   & 43\%   & \textbf{50\%}      \\
wave            & 7\%    & \textbf{20\%}   & \textbf{20\%}      \\\hline\hline
Average			& 48\%	 & \textbf{58\%}   & 55\%      \\
\end{tabular}
\label{table:4}
\end{table}


\noindent\paragraph{Benefits of Three-stream Model:}
In Tables \ref{table:2} and \ref{table:3}, we also evaluate our three-stream network against the original two-stream framework~\cite{simonyan2014very}. It can be seen that the performance of DF + RGB is similar to Stack(S) of optical flow + RGB (which is the standard two-stream model) on both UCF101 and HMDB51 datasets. However, interestingly, we find that, after combining stacked optical flow with DF+RGB, the accuracy improves by 3\% on HMDB51 and 4\% on UCF101, which shows that our new representation carries complementary information (such as long-range dynamics) that is absent previously. 

\begin{table}[]
\centering
\caption{Accuracy comparison between AlexNet and VGG-16 on HMDB51 split 1.}
\begin{tabular}{lll}
Method              & Alex net                     & VGG-16 \\ \hline
Static RGB          & 40.50\%\cite{simonyan2014two}& 47.12\% \\
Dynamic Flow        & 43.69\%       			   & 48.75\% \\
Dynamic Image       & 40.88\%                      & 44.74\% \\
Dynamic Flow + RGB  &\textbf{53.75\%}              &\textbf{58.30\%}\\
Dynamic Image + RGB & 45.21\%                      & 47.96\% \\\hline
\end{tabular}
\label{table:5}
\end{table}

\begin{table}[]
\centering
\small
\caption{Accuracy comparison between AlexNet and VGG-16 on UCF101 split 1.}
\begin{tabular}{lll}
Method                           & Alex net        & VGG-16       \\ \hline
Static RGB                       & 71.20\%\cite{wang2015action}& 80.00\% \\
Dynamic Flow                     & 75.36\%         & 78.00\% \\
Dynamic Flow+RGB                 &\textbf{84.25\%} &\textbf{87.63\%}\\
Dynamic Flow+RGB+(S)Optical flow &\textbf{88.65\%} &\textbf{90.30\%}\\\hline
\end{tabular}
\label{table:6}
\end{table}

\noindent\paragraph{Comparisons between CNN architectures:} To further validate and understand the behavior of the three-stream model, we repeated the experiment in the last section using an Alexnet architecture. As is shown in Table \ref{table:5} and \ref{table:6}, the accuracy of VGG-16 is seen to be 2\%-7\% higher than for Alex net, which is expected. We also find that the performance of DF, DF + RGB and DF + RGB + stacked optical flow show the same trend in both Alex net and VGG-16 networks.

\begin{table}[]
\centering
\small
\caption{Accuracy comparison on HMDB51 split 1 using VGG-16 after combining with IDT-FV.}
\begin{tabular}{ll|l}
Method                            & +IDT-FV     &Original\\ \hline
Static RGB                        & 61.89\%     &47.06\%\\
Dynamic Flow					  & 58.30\%     &48.75\%\\
Dynamic Image                     & 47.96\%     &44.74\%\\
Dynamic Image+RGB                 & 65.44\%     &47.96\%\\
Dynamic Flow+RGB                  &{67.35\%}&{58.30\%}\\
Dynamic Flow+RGB+(S)Optical flow  &\textbf{67.48\%}&\textbf{61.70\%}\\
Dynamic Image+RGB+(S)Optical flow & 64.13\%     &54.40\%\\\hline
\end{tabular}

\label{table:7}
\end{table}

\begin{table}[]
\centering
\small
\caption{Accuracy comparisons on UCF101 split 1 using VGG-16 after combining with IDT-FV.}
\begin{tabular}{ll|l}
Method                            & +IDT-FV     &Original\\ \hline
Dynamic Flow+RGB                  &\textbf{89.20\%}&{87.63\%}\\
Dynamic Flow+RGB+(S)Optical flow  &\textbf{91.10\%}&{90.30\%}\\\hline
\end{tabular}

\label{table:8}
\end{table}

\noindent\paragraph{Benefits from Hand-crafted Features:}
In Table \ref{table:7} and \ref{table:8}, we evaluate the performance of three-stream network and its combination with IDT-FV~\cite{wang2013dense}. On HMDB51, after applying IDT-FV, the accuracy of each method improves by 2\% to 10\%. While, using this combination also improves the accuracy for DI + RGB, this improvement is significantly inferior to DF+RGB. On the UCF101 dataset, IDT-FV improves performance by 1-- 2\%. 

\noindent\paragraph{Comparisons to the State of the Art:}
In Table~\ref{table:9}, we compare our method against the state-of-the-art results on HMDB51 and UCF101 datasets. For this comparison, we use the VGG-16 model trained on dynamic flow images, combined with static RGB, a stack of 10 optical flow frames, and IDT-FV features. The results are averaged over three splits as is the standard protocol. For both the datasets, we find that our three stream model with dynamic flow images outperforms the best results using dynamic image networks~\cite{bilen2016dynamic}. For example, on HMDB51, our results by replacing the dynamic image with dynamic flow leads to a 2\% improvement. We also notice a performance boost against the recent hierarchical variant~\cite{fernando2016discriminative} of the dynamic images (66.9\% against 67.35\%) that recursively summarizes such images from video sub-sequences.

\begin{figure}[ht]
\centering
\small
	\includegraphics[width=0.95\linewidth,trim={0cm 0cm 0cm 0.5cm}]{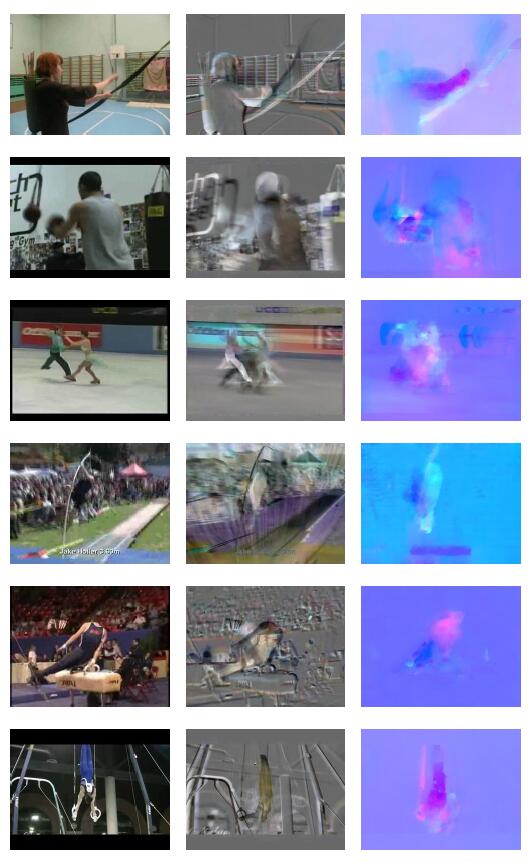}
   \caption{Left to right: Qualitative results of RGB frames, dynamic images, and dynamic flow on UCF101 dataset. }
\label{fig:4}
\end{figure}
\begin{figure}[ht]
\centering
\small
	\includegraphics[width=0.95\linewidth,trim={0cm 0cm 0cm 0.5cm}]{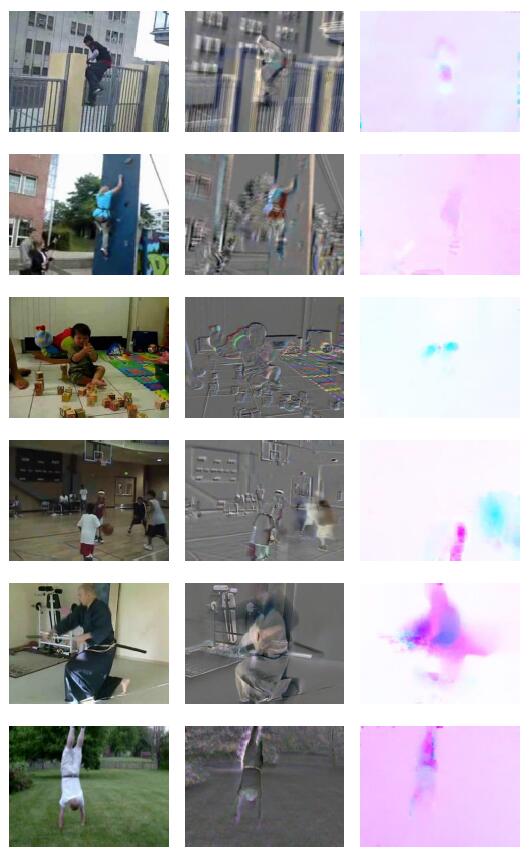}
   \caption{Left to right: Qualitative results of RGB frames, dynamic images, and dynamic flow on HMDB51 dataset.}
\label{fig:5}
\end{figure}

Compared to other state-of-the-art methods, our results are very competitive. Notably, while the accuracy of two-stream fusion \cite{feichtenhofer2016convolutional} is slightly higher than ours, our CNN architecture is significantly simpler. In Figures~\ref{fig:4} and~\ref{fig:5}, we provide qualitative comparisons between dynamic flow and dynamic images.


\begin{table}[]
\centering
\small
\caption{Classification accuracy against the state of the art on HMDB51 and UCF101 datasets averaged over three splits.}
\begin{tabular}{lcc}
Method & \small{HMDB51} &\small{UCF101}  \\\hline
Two-stream \cite{simonyan2014two}                       & 59.40\%   & 88.00\%\\

Two-stream SR-CNNs \cite{wang2016two}                       &--  & 92.60\%\\
Very Deep Two-stream Fusion \cite{feichtenhofer2016convolutional} & \textbf{69.20\%} & \textbf{93.50\%}  \\
LSTM-MSD \cite{li2016action}                              &63.57\% & 90.80\%\\
IDT-FV \cite{wang2013action}                            & 57.20\% & 86.00\%\\
IDT-HFV \cite{peng2016bag}                              & 61.10\% & 87.90\%\\
TDD+IDT-FV \cite{wang2015action}                           & 65.90\% &91.50\%\\
C3D + IDT-FV~\cite{tran2015learning} & -- & 90.40\% \\
Dynamic Image + RGB + IDT-FV \cite{bilen2016dynamic}                   & 65.20\% & 89.10\% \\
Hierarchical Rank Pooling~\cite{Fernando2016a} & 66.90\% & 91.40\%\\
\hline
Ours &&\\
Dyn. Flow + RGB + IDT-FV                      &67.19\% & {89.41\%} \\
Dyn. Flow+RGB+(S)Op.Flow+IDT-FV             & {67.35\%} &{91.32\%}\\
\hline
\end{tabular}
\label{table:9}
\end{table}

\section{Conclusions}
In this paper, we presented a novel video representation -- dynamic flow-- that summarizes a set of consecutive optical flow frames in a video sequence as a single two-channel image. We showed that our representation can compactly capture the long-range dynamics of actions. Considering this representation as an additional CNN input cue, we proposed a novel three-stream CNN architecture that incorporates single RGB frames (for action context), stack of flow images (for local action dynamics), and our novel dynamic flow stream (for long range action evolution). Experiments were provided on standard benchmark datasets (HMDB51 and UCF101) and clearly demonstrate that our method is promising in comparison to the state of the art. More importantly, our experimental results reveal that our representation captures the action dynamics more robustly than the recent dynamic image algorithm and provides complimentary information (long-range) compared to the traditional stack of flow frames.

\noindent\paragraph{{Acknowledgements:}}{AC is supported by the Australian Research Council (ARC) through the Centre of Excellence for Robotic Vision (CE140100016).

{\small
\bibliographystyle{ieee}
\bibliography{egbib}
}

\end{document}